\documentclass[preprint,12pt]{elsarticle}

\usepackage{amssymb} 

\usepackage{xspace}
\makeatletter
\DeclareRobustCommand\onedot{\futurelet\@let@token\@onedot}
\def\@onedot{\ifx\@let@token.\else.\null\fi\xspace}
\def\eg{\emph{e.g}\onedot} 
\def\ie{\emph{i.e}\onedot}

\makeatother
\usepackage{amsmath}
\usepackage{adjustbox}
\usepackage{array}
\usepackage{bbm}
\usepackage[nolist,nohyperlinks]{acronym}
\usepackage{xcolor}
\usepackage{adjustbox}
\usepackage{rotating}
\usepackage{makecell}
\usepackage{multirow}
\usepackage{arydshln}
\usepackage{url}
\usepackage[hidelinks]{hyperref}

\begin{acronym}
\acro{BoVW}{Bags of Visual Words}
\acro{VO}{Visual Odometry}
\acro{SLAM}{Simultaneous Localization and Mapping}
\acro{BA}{Bundle Adjustment}
\acro{PGO}{Pose-Graph Optimization}
\acro{HiRDLS}{High-Resolution Dynamics Limb Sounder}
\acro{TES}{Tropospheric Emission Sounder}
\end{acronym}

\makeatletter
\def\ps@pprintTitle{%
\let\@oddhead\@empty
\let\@evenhead\@empty
\def\@oddfoot{}%
\let\@evenfoot\@oddfoot}
\makeatother

\begin{document}

\begin{frontmatter}



\title{H-SLAM: Hybrid Direct-Indirect Visual SLAM}




\author{Georges Younes\textsuperscript{1,2}}
\author{Douaa Khalil\textsuperscript{1}}
\author{John Zelek\textsuperscript{2}}
\author{Daniel Asmar\textsuperscript{1}}
\address{\textsuperscript{1}Vision and Robotics Lab, American University of Beirut, Beirut, Lebanon}
\address{\textsuperscript{2}Systems Design Department, University of Waterloo, Waterloo, Ont. Canada}

\begin{abstract}
The recent success of hybrid methods in monocular odometry has led to many attempts to generalize the performance gains to hybrid monocular SLAM. However, most attempts fall short in several respects, with the most prominent issue being the need for two different map representations (local and global maps), with each requiring  different, computationally expensive, and often redundant processes to maintain. Moreover, these maps tend to drift with respect to each other, resulting in contradicting pose and scene estimates, and leading to catastrophic failure.
In this paper, we propose a novel approach that makes use of descriptor sharing to generate a single inverse depth scene representation. This representation can be used locally, queried globally to perform loop closure, and has the ability to re-activate previously observed map points after redundant points are marginalized from the local map, eliminating the need for separate and redundant map maintenance processes. The maps generated by our method exhibit no drift between each other, and can be computed at a fraction of the computational cost and memory footprint required by other monocular SLAM systems. 
Despite the reduced resource requirements, the proposed approach maintains its robustness and accuracy,  delivering performance comparable to state-of-the-art SLAM methods (\eg, LDSO, ORB-SLAM3) on the majority of sequences from well-known datasets like EuRoC, KITTI, and TUM VI. The source code is available at: \url{https://github.com/AUBVRL/fslam_ros_docker}.
\end{abstract}



\begin{keyword}
Monocular \sep SLAM \sep Hybrid \sep Direct \sep Indirect methods.


\end{keyword}

\end{frontmatter}


\section{Introduction}
\label{sec:Introduction}

Recently, hybrid methods have emerged as \ac{VO} solutions that harness the strengths of both direct and indirect (feature-based) formulations, to yield superior performance compared to each individual formulation.
For instance, the sub-pixel accuracy and robustness to texture-deprived environments of direct methods, combined with the resilience of indirect methods to large motions, have been shown to outperform current state-of-the-art methods in both direct and indirect \ac{VO} methods \cite{younes_2019_iros}.

The success of hybrid \ac{VO} systems led to many methods that aim to extend this symbiotic relationship from an odometry formulation to a \ac{SLAM} one. However, most attempts fall short in one or several aspects. For example, LDSO \cite{Gao_2018_iros} makes use of indirect features to efficiently query global keyframes for loop closure, but does not utilize these features for neither tracking nor mapping. On the other hand, \cite{lee_2019_ral} deploy a cascaded approach where a direct odometry is run at frame rate. The resulting output is subsequently fed into an indirect framework for additional processing and global mapping. This approach results in two disjoint map representations, having completely different and redundant map maintenance and optimization processes, which can lead to decreased accuracy due to the potential drift between the two maps. 

More recently, \cite{luo_2021_ral} proposed a double window approach that leverages both direct and indirect formulations in a \ac{SLAM} framework, where three sequential optimizations, namely, photometric \ac{BA}, double window optimization, and structure-only optimization, are used to maintain the local and global maps separately. While this work is the most promising approach for implementing a hybrid SLAM system to date, it still processes and maintains two separate map representations.

Having two distinct map representations introduces several limitations into a \ac{SLAM} system; for example, the two representations can drift with respect to each other, yielding contradictory pose and scene structure estimates, which can in turn lead to catastrophic failure. While this is typically mitigated by separate, often redundant, and computationally expensive map maintenance processes, such catastrophic failure is inevitable due to the sub-optimal nature of these maintenance methods. In contrast, our proposed method uses the same process to build, maintain, and update both local and global map representations. In essence, they are the same map representation. The only difference between a local and global map point is that a global map point is not included in the local photometric bundle adjustment, but is added again as soon as it is re-observed. This allows us to build a coherent and re-usable \ac{SLAM} map that remains immune to drift, and does not require a separate and computationally expensive maintenance processes.

On another note, most of the aforementioned \ac{SLAM} systems employ \ac{PGO} for loop closure. While \ac{PGO} is computationally efficient, it is a sub-optimal optimization that neglects all 3D observations.
In contrast, we propose to perform a full global inverse depth \ac{BA} on the entire map, taking into account the 3D observation constraints. While this is typically not possible in systems similar to \cite{Gao_2018_iros}, it is possible in ours as we keep track of the connectivity information of the map points across keyframes. 

Moreover, it is common in most \ac{PGO} implementations to utilize only one type of constraint. Typically, either co-visibility constraints (as described in \cite{mur-artal_2015_TRO}) or pose-pose constraints (as described in \cite{Gao_2018_iros}) are employed. Each representation offers its own advantages and disadvantages. It is rare for systems to use both types concurrently, resulting in compromised performance when re-observing previous scenes with methods relying on pose-pose constraints, and facing catastrophic failures when traversing feature-deprived environments with methods relying on co-visibility constraints. To tackle this challenge, we propose a hybrid connectivity graph that keeps track of both pose-pose and co-visibility constraints. By adopting this approach, our system effectively handles temporal constraints in feature deprived environments, while maintaining the ability to establish long range constraints when revisiting past scenes.

To address all of the aforementioned issues, we leverage our descriptor sharing formulation presented in \cite{younes_2019_iros} and extend its capabilities to maintain a global map. The proposed hybrid SLAM system performs a joint and tightly-coupled (direct-indirect) optimization for tracking. It then uses the same inverse depth parametrization for mapping, where global map points are excluded from the local photometric \ac{BA} and reintegrated whenever they are re-observed (\textit{i.e.}, the same map representation can be concurrently used locally and globally).
Furthermore, we capitalize on the advantages of this joint formulation to generate hybrid connectivity graphs without incurring any additional computational cost. These graphs contain both co-visibility and pose-pose constraints, which can in turn be used either for \ac{PGO} or to produce optimal results using a full inverse-depth \ac{BA} over the keyframe poses and associated landmarks. 

\noindent The key aspects of the proposed system are:

\begin{enumerate}
\item The ability to leverage the advantages of both direct and indirect formulations to achieve a robust camera pose estimation. The achieved accuracy is comparable to existing state-of-the-art methods, and in some cases, surpasses them in both direct and indirect monocular \ac{SLAM}/Odometry systems.

\item The ability to compute both local and global map representations at a fraction of the time required by indirect methods to build and maintain a global map alone, while introducing a mere increase of 13 ms per keyframe\footnote{Reported timings in the paper are subject to the hardware used. More in Sec. \ref{sec:compute_cost}.} when compared to strictly local direct odometry systems such as DSO.

\item The ability to identify and re-use global map points efficiently.

\item A reduced memory consumption, where the proposed unified global-local map requires less than half of the memory typically consumed by other \ac{SLAM} methods.  

\item The proposed approach keeps track of both pose-pose and co-visibility constraints, utilizing them strategically during loop closure for added robustness in feature-deprived environments. Additionally, our method enables the ability to re-populate the co-visibility constrains after loop closure is performed, allowing for optimal subsequent optimization processes such as full inverse depth \ac{BA} and efficient re-use of global map points.

\end{enumerate}
\section{Related work}
\label{sec:RelatedWork}
Before delving into our proposed solution, it is important to highlight the various limitations of current Visual \ac{SLAM} (VSLAM) systems as a prelude to the significance of our contributions, particularly in the areas of local and global mapping, as well as loop closure.

\subsection{Local and global mapping: a parametrization issue}
\label{sec:local_global_mapping}
Over the years, the research community has adopted several formulations to represent local and global maps, grounded by the data type (direct vs indirect) available to the localization and mapping systems. For example, most indirect methods such as \cite{klein_2007_ISMAR, mur-artal_2015_TRO} employ traditional multi-view geometry to perform triangulation across large baselines. This is possible because their features are resilient to viewpoint changes. The triangulation method entails representing the map as an $(X,Y,Z)$ point cloud. However, it suffers from the instability of multi-view geometry under small parallax. In contrast, direct methods like \cite{forster_2014_ICRA, engel_2014_ECCV, engel_2016_ARXIV} require small motion between frames for brightness constancy to be preserved, making traditional multi-view triangulation methods unsuitable. Instead, the map is updated via sequential small-baseline observations using a filter-based formulation such as \cite{forster_2014_ICRA} and \cite{engel_2013_ICCV}. This triangulation method entails an inverse depth parametrization $(u,v, \frac{1}{d})$ \cite{civera_2008_TRO} that can transition smoothly from infinity under no parallax to a triangulated depth value as more observations come through.

Converting between these two parametrizations is fairly trivial. However, the associated triangulation methods differ, making it difficult to use both concurrently. For example, one method requires the feature match locations across all observed images, while the other requires a consistent depth variance estimate across observations. Although this is not an issue when a system adopts a single representation, it becomes problematic in hybrid methods such as \cite{lee_2019_ral, younes_2018_visapp, luo_2021_ral} where the two representations are used. The non-interchangeability of the triangulating methods forces these systems to adopt separate map maintenance processes (\textit{e.g.} structure-only optimization) to ensure no drift occurs between both representations. 

In our previous work \cite{younes_2019_iros}, we circumvented the need for several map maintenance processes 
by operating within a pure odometry system. Redundant map points that were marginalized out of the local window were completely removed and not subsequently used in a global framework. In this paper, we rely on the same unified formulation, but extend its capabilities to maintain both representations using a single process. This means that instead of having separate maps, we employ a unified map that can be used interchangeably as either an $(X,Y,Z)$ point cloud or as a $(u,v,\frac{1}{d})$ inverse depth parametrization.


\subsection{Loop closure essentials}
\label{sec:loopClos}

As an odometry system integrates new measurements with old ones, accumulated numerical errors can grow unbounded, resulting in a \textit{drift} between the actual camera trajectory, the 3D scene, and their corresponding system estimates. In fact, a monocular odometry system can drift along the scale, rotation, and translation axes, leading to an ill-representation of the environment and eventual failure.

Loop closure is a mechanism used to identify and correct drift. It can be split into three main tasks: (1) loop closure detection, (2) error estimation, and (3) correction.\\

\noindent\textbf{(1) Loop closure detection} is the process of flagging a candidate keyframe from a set of previously observed keyframes that exhibit similar \textit{visual cues} to the most recent keyframe (\ie, the camera has returned to a previously observed scene). 
While most \ac{SLAM} systems perform loop closure by following the above three tasks, the implementation differs according to the type of information at their disposal. For example, low level features (\eg, patches of pixels) are susceptible to viewpoint and illumination changes, and are inefficient at encoding and querying an image. Therefore, most direct odometry systems discard them once they go out of view (\eg DSO \cite{engel_2016_ARXIV}). To perform loop closure, direct systems require auxiliary features; for example, LSD SLAM \cite{engel_2014_ECCV} extracts STAR features \cite{agrawal_2008_eccv} from keyframes and associates them with SURF descriptors \cite{bay_2008_CVIU}, which are in turn parsed through OpenFABMAP \cite{glover_2012_ICRA} (an independent appearance-only SLAM system) to detect loop closure candidates. Similarly, LDSO \cite{Gao_2018_iros} extracts corner features \cite{shi_1994_CVPR} with their associated ORB descriptors \cite{rublee_2011_ICCV}, then uses them to encode and query the keyframes for loop closure candidate detection using a \ac{BoVW} model \cite{lopez_2012_TRO}.

In contrast, ORB-SLAM (1, 2 and 3) \cite{mur-artal_2015_TRO, Mur-Artal_Tardos_2017, campos_2021_tro} does not require separate processes; the same features used for odometry (FAST features \cite{rosten_2006_ECCV} and ORB descriptors \cite{rublee_2011_ICCV}) are parsed in a \ac{BoVW} \cite{lopez_2012_TRO} to compute a global keyframe descriptor, which is then used to query a database of previously added keyframes for potential loop closure candidates.\\

\noindent\textbf{(2) Error estimation}: once a candidate keyframe is detected, its corresponding map point measurements are matched to the most recently added 3D map points, then a similarity transform $T\in$ \textbf{Sim(3) := $\left[s\mathbf{R} | \mathbf{t}\right]$} that minimizes the error between the two 3D sets is computed. This process is slightly different between the two commonly used loop closure methods, but can have a significant impact on the resulting accuracy.
In particular, ORB-SLAM establishes 3D-3D map point matches between the loop-ends, which are then used in a RANSAC implementation of \cite{horn_1987_JOSA} to recover a similarity transform. The new similarity is used to search for additional map point matches across other keyframes connected to both loop ends, resulting in a relatively large set of 3D map point matches. The entire set is eventually employed to refine the estimation of the similarity transformation.

In contrast, LDSO only establishes 3D map point matches between the currently active keyframe and one loop candidate keyframe. The loop candidate keyframe is also found through \ac{BoVW}; however, it has one single requirement, which is to be outside the currently active window. As a consequence, the number of used matches is considerably lower compared to ORB-SLAM, often resulting in a non-reliable similarity transformation. Moreover, since LDSO only requires a loop closure candidate to exist outside the current active window, it often performs consecutive loop closures whenever the camera transitions away from a scene and returns to it few seconds later. These recurrent loop closures, along with the non-reliable similarity transforms, can introduce significant errors into the 3D map.\\


\noindent\textbf{(3) Path correction.} The error estimation process corrects the loop-end keyframes; however, it does not correct the accumulated drift throughout the entire path. For that, a \ac{SLAM} system must keep track of the connectivity information between keyframes along the traversed path in the form of a graph, to subsequently correct the entire trajectory. Most systems employ a \ac{PGO}, which is a sub-optimal optimization that distributes the accumulated error along the path by considering frame-to-frame constraints exclusively, and disregarding the 3D observations.

\begin{sloppypar}
To be able to perform \ac{PGO}, a notion of connectivity between the keyframes must be established. To that end, ORB-SLAM employs several representations of such connections. Notably, the \textbf{co-visibility} graph is a byproduct of ORB-SLAM's feature matching on every keyframe, where a connectivity edge is inserted between keyframes that observe the same 3D map points. ORB-SLAM also uses a \textbf{spanning tree} graph, made from a minimal number of connections, where each keyframe is connected to its reference keyframe and to one child keyframe. Finally, ORB-SLAM also keeps track of an \textbf{essential graph}, which contains the spanning tree and all edges between keyframes that share more than $100$ feature matches. Note that the spanning tree $\subseteq$ essential graph $\subseteq$ co-visibility graph; and while the full co-visibility graph can be used for \ac{PGO}, ORB-SLAM relies on the essential graph and cites the computational efficiency as a reason for this choice, since the former might introduce a large number of constraints.
\end{sloppypar}

Unlike ORB-SLAM, which establishes connectivity graphs by finding feature matches between keyframes, LDSO lacks access to such information, as it does not perform feature matching between keyframes nor does it keep track of feature matches. Instead, it considers all currently active keyframes in the local window to be connected, and accordingly adds an edge between them in its connectivity graph. While this works well in feature-deprived environments, where not enough feature matches can be established, it has several drawbacks when compared to its ORB-SLAM counterpart.
Specifically, whenever a loop closure takes place in ORB-SLAM, new connections between keyframes that were previously disconnected due to drift are updated based on their mutually observed map points; such update is not possible within LDSO's model.

This is where our proposed hybrid connectivity graph plays an important role; by maintaining both connectivity information, a hybrid graph enables the addition of connections between keyframes based on their shared 3D map points. This is particularly useful in situations involving loop closure or when previous scenes are re-observed. Additionally, the hybrid graph maintains the capability to establish connections in feature-deprived environments by considering the temporal proximity of the added keyframes. Furthermore, adding both types of information allows for more optimal optimization methods such as full \ac{BA} on both the path and reconstructed scene; albeit, at an extra computational cost.

\subsection{Global map memory management}
For proper operation, direct systems typically add a relatively large number of keyframes per second. While this is generally not an issue for pure odometry methods, where the memory consumption remains constant, the unbounded memory growth becomes an issue for \ac{SLAM} systems that maintain and re-use a global map representation.  To preserve a reasonable memory consumption and keep the compute-time low, ORB-SLAM invokes a keyframe culling strategy that removes keyframes whose 90\% of map points are visible in other keyframes. This, however, has a negative impact on the final result's accuracy since culled keyframes are completely removed from the system, whereas their poses could have been used to better constrain the estimated trajectory.

On the other hand, while LDSO does not utilize its global map for pose estimation and mapping, it still needs to store all keyframes in memory along with their associated indirect features and depth estimates in order to perform loop closure. Since it also does not perform feature matching to detect redundant keyframes, it cannot invoke a keyframe culling strategy, resulting in a relatively large memory consumption.

Finally, despite the fact that DSM \cite{Zubizarreta_2020_tro} does not have a loop closure module, it blurs the line between what can be classified as an odometry system \textit{vs.} a \ac{SLAM} system: it is a direct method that keeps track of a global map, and attempts to re-use it for both pose estimation and mapping, without performing loop closure. However, this implies the need to store a large number of information per keyframe, and retrieve them whenever a new keyframe is added. As a consequence, the resulting global map requires a relatively significant amount of memory, and the computational cost increases with the addition of each new keyframe. DSM addresses these challenges by adding fewer keyframes than other direct methods.

We mitigate the issues of frequent keyframe addition, along with its associated memory growth and increased map querying time, by using descriptor sharing. In our formulation, a map point can have several representations such as ORB descriptors, patch of pixels, and more. This enables efficient querying of landmarks from both the local and global map. We also maintain connectivity information between keyframes, which allows us to perform keyframe culling. While keyframes are removed, their poses are retained within the pose-pose constraints graph, enabling us to refine the trajectory in the future. These memory management routines allow us to maintain both local and global representations of a scene at a smaller memory footprint compared to state-of-the-art direct, indirect, and other hybrid methods.
\section{Proposed system}
\label{sec:ProposedSystem}
\begin{figure*}[!ht]
    \centering
    \includegraphics[width=\textwidth]{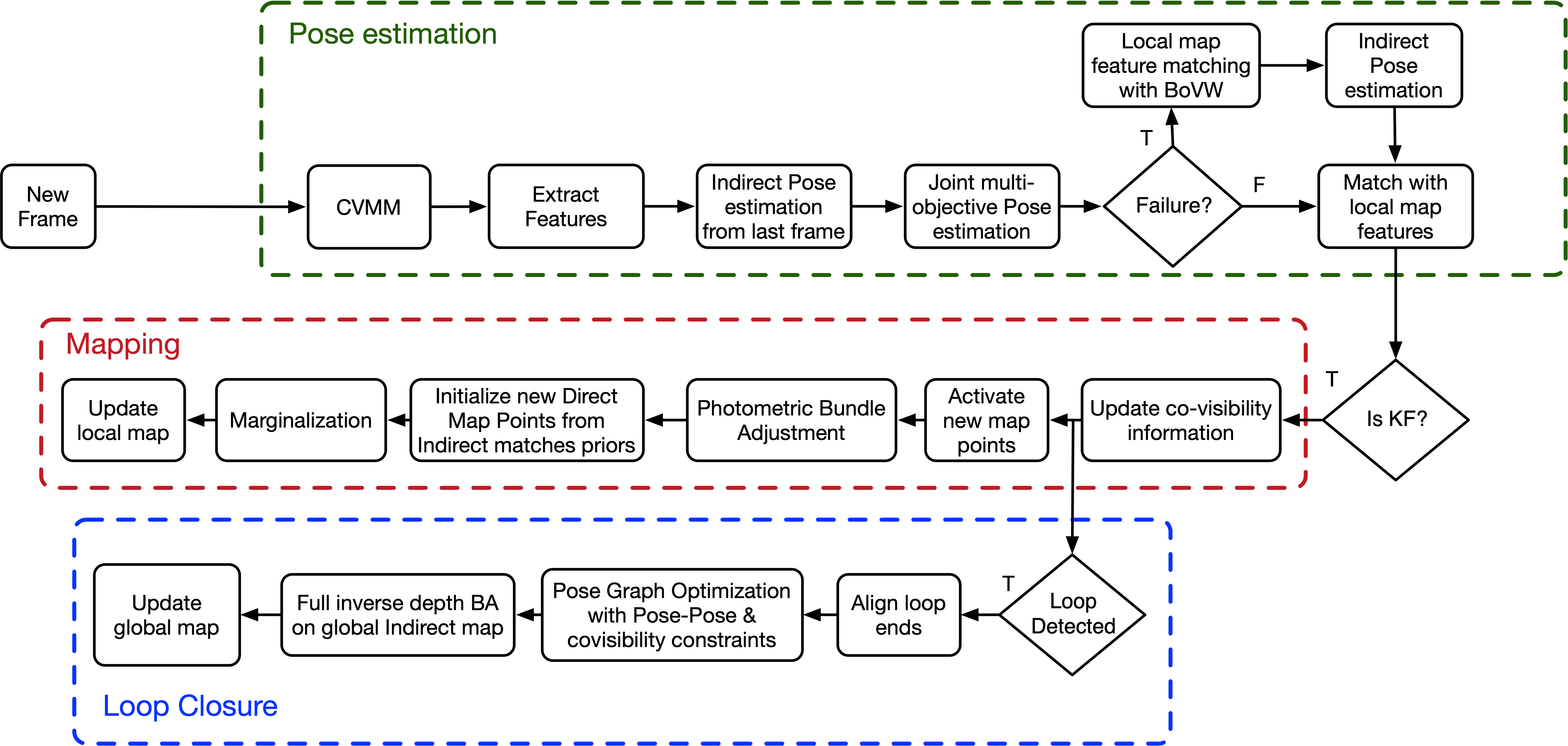}
    \caption{Proposed system architecture made of three parallel threads, namely joint pose estimation thread, mapping thread and a loop closure thread.}
    \label{fig:proposed_architecture}
\end{figure*}

The proposed system architecture (shown in Fig. \ref{fig:proposed_architecture}) follows a fairly standard Visual SLAM architecture with three parallel threads, namely, pose estimation, mapping, and loop closure. However, as will be detailed in this section, we introduce several key modifications to generate both local and global maps concurrently, \ie, with no separate map representations. This approach allows us to efficiently perform the joint pose optimization, to maintain the current moving window and the set of redundant features that were marginalized within the same process, and to perform loop closure all within the same framework. 
The key concept that enables us to achieve this is descriptor sharing, which was initially proposed in our previous work \cite{younes_2019_iros}.

\subsection{Descriptor sharing}
\begin{figure}
    \centering
    \includegraphics[width=0.6\linewidth]{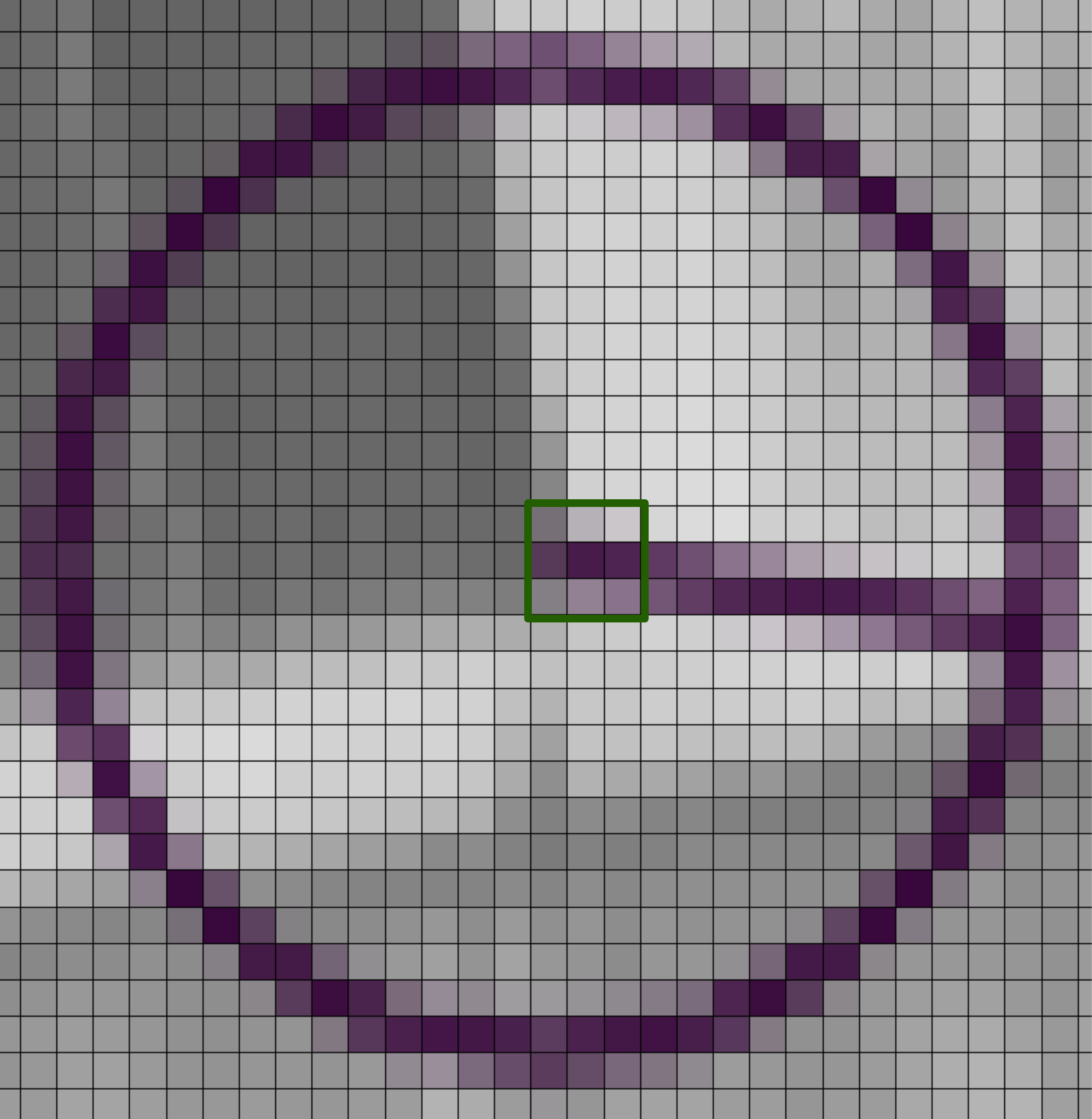}
    \caption{Descriptor sharing where one feature can have several descriptors; in this case it has an ORB descriptor represented with the purple circle, and a patch of pixels descriptor represented with the green square.}
    \label{fig:shared_desc}
\end{figure}

Descriptor sharing is the idea of associating several types of descriptors with the same feature. To illustrate, one could detect corners and simultaneously associate them with both an ORB descriptor \cite{rublee_2011_ICCV} and a patch of pixels as shown in Fig. \ref{fig:shared_desc}. This enables the use of each descriptor in its favorable conditions to perform various SLAM tasks. For instance, the patch of pixels can be employed for low-parallax triangulation, while the ORB descriptor can be used to perform large-baseline feature-matching. This matching can be leveraged for tasks such as pose estimation, constructing connectivity graphs, maintaining a global map for landmark re-use, and more.

Therefore, within the context of this paper, it is crucial to establish clear definitions of the different features, their corresponding residuals, and their respective uses, as well as what we mean by \textit{local} and \textit{global} maps. This will ensure a common understanding before proceeding with further details. \\

\textbf{Features:} in this work, a blend of feature types is extracted: first, we detect corners using the FAST detector \cite{rosten_2006_ECCV}, then augment them with pixel locations whose gradient surpasses a dynamically adjusted cutoff threshold, as described in \cite{engel_2016_ARXIV}. There are two primary reasons for using a combination of features.
Firstly, FAST corners are repeatable and stable; in contrast, pixel locations with high gradients (gradient-based features) are typically detected along edges, which makes them unstable as they tend to drift along the edge directions, thereby reducing the overall VSLAM performance. Secondly, a texture-deprived environment can cause a significant decrease in detected corners and may lead to tracking failure. In contrast, gradient-based features can be abundantly extracted along any gradient in the image. Having some information in such scenarios is better than having no information.\\
Once the two types of features are extracted, they are treated equally in the subsequent steps, \ie, they are associated with both ORB descriptors and patches of pixels. The ORB descriptors will be used to do feature matching, establish a geometric re-projection error, maintain a global co-visibility graph among keyframes, and perform loop closures. On the other hand, the patch of pixels will contribute towards computing the photometric residuals, estimating the depth of the features as described in \cite{engel_2013_ICCV}, and performing the photometric \ac{BA}.

\textbf{Local vs. global maps:} while we adopt the conventional naming of local and global maps used in hybrid approaches, our implementation introduces a slight but fundamental difference. We represent landmarks in both definitions using the same inverse depth parametrization, effectively allowing us to re-activate the global landmarks in the local window when re-observed. This is not feasible in previous hybrid approaches as they convert the marginalized redundant map points to an $(X, Y, Z)$ representation, consequently requiring separate map maintenance processes to mitigate the drift between the two representations.

\subsection{Pose estimation}
We estimate the camera pose in two sequential steps: (1) we perform feature matching and relative pose estimation with respect to the last frame; (2) we proceed with a joint multi-objective pose optimization over both photometric and geometric residuals.

While we could solely rely on the joint multi-objective optimization, we have found that performing step (1) beforehand can help establish more reliable (\ie, less outlier) feature matches in the subsequent joint optimization. This is primarily because we can narrow down the search window for matches when matching sequential frames. With fewer outliers, 
the overall computational cost remains fairly similar to using the joint optimization alone. In order to address situations where an insufficient number of matches are found in the previous steps, we implemented a failure recovery mechanism using \ac{BoVW}. This process is summarized in the green block of Fig. \ref{fig:proposed_architecture}.\\

\noindent\paragraph{Joint multi-objective pose optimization}
We compute the camera pose by concurrently minimizing photometric and geometric residuals. Since both types of descriptors have different but complementary properties, a utility function is introduced to analyze the current observations and accordingly modify the weights of each residual type as the optimization progresses. This scalarized multi-objective optimization is summarized as:
\begin{equation}
\label{eq:ScalarizedRed}
\underset{\xi}{\operatorname{argmin}}\:\mathbf{e}(\xi)=\underset{\xi}{\operatorname{argmin}}\left[\frac{\|\mathbf{e}_p(\xi)\|_\gamma}{n_p\sigma_p^2}+K\frac{\|\mathbf{e}_g(\xi)\|_\gamma}{n_g\sigma_g^2}\right],
\end{equation}
where $K$ is the utility function's output, $n$ is the count of each feature type, $\sigma^2$ is the residual's variance\footnote{Residual balancing is a very important aspect of multi-objective optimization that is often neglected by VSLAM practitioners, leading to fallacies such as the joining of pose-pose constraints with geometric re-projection errors. In this work we balance the residuals by normalizing against their variance and number of measurements.}, $\parallel\cdot\parallel_\gamma$ is the Huber norm, and the energy per feature type is defined as:
\begin{equation}
\label{eq:errorval}
 \mathbf{e}_{p}(\xi)=(\mathbf{r}^TW\mathbf{r})_p,\: and\; \mathbf{e}_{g}(\xi)=(\mathbf{r}^TW\mathbf{r})_g,
\end{equation}
$\mathbf{r}$ is the vector of stacked residuals per feature type, \ie, $r_p$ represents the photometric residuals (pixel intensity differences) and $r_g$ represents the geometric re-projection residuals (pixel distance differences). Finally, 
\begin{equation}
\label{eq:var_weight}
W=\mathbbm{1}\left(\frac{\frac{1}{\sigma^2_d}}{max\left(\frac{1}{\sigma_d^2}\right) }\right),
\end{equation} 
is a weight matrix that dampens the effect of landmarks with large depth variance on the pose estimation process.

\noindent\paragraph{Logistic utility function}
The logistic utility function provides a mechanism to steer the multi-objective optimization as it progresses, allowing us to incorporate prior information on the behaviour of the different descriptor types. For example, pixel-based residuals have a small convergence basin \cite{younes_2018_visapp, engel_2014_ECCV}, whereas geometry-based residuals are better behaved when starting the optimization from a relatively far initializing point. As such, the utility function gives higher weights to the geometric residuals in the early stages of the optimization then gradually shifts the weight towards the pixel-based residuals. Similarly, geometric residuals are prone to outliers in texture-deprived environments and under motion blur. The proposed logistic utility function decreases the influence of the geometry-based residuals when the number of feature matches is low. Both of these effects are captured by:
\begin{equation}
\label{eq:Utility}
K=\frac{5e^{-2l}}{1+e^{\frac{30-N_g}{4}}},
\end{equation} 
where $l$ is the pyramid level at which the optimization is taking place, and $N_g$ is the number of current inlier geometric matches.


\subsection{Mapping}
\label{sec:proposed_mapping}
\noindent\paragraph{Local map (Fig. \ref{fig:proposed_architecture} - red block)} This map follows a similar definition of direct \ac{VO} systems \cite{engel_2016_ARXIV}; it consists of a moving window that contains 7 keyframes. As new keyframes are added, the oldest keyframes are marginalized, ensuring a fixed-size local map as described in \cite{leutenegger_2015_ijrr}. The map also contains a set of landmarks (features) associated with both patches of pixels and ORB descriptors, and whose depth estimate can be modified within a local photometric \ac{BA}. Since keyframe addition and removal is performed through marginalization, the local map also has prior factors that encode the probability distribution of the remaining states in the local map given the marginalized redundant data. This process makes local maps difficult to modify as any edits or subsequent post-processing like loop closures would render the prior factors meaningless, and introduce significant errors into the local \ac{BA}.

In our previous work \cite{younes_2019_iros},
we extended the local map beyond the currently active set of features to include recently marginalized landmarks that can still be matched to the latest keyframe using their ORB descriptors. This allows recently marginalized landmarks to contribute towards the pose estimation. However, since these landmarks have a different paramterization than their local counterparts, we could not re-activate them within the local window, nor could we maintain them for future re-use. Instead, keyframes in the extended local map along with their features were completely dropped whenever all of the corresponding features were no longer observed in the latest keyframe. 

In this work, we leverage this extended set of keyframes and landmarks beyond assisting the local active map, to build a global and queryable map that enables loop closure and allows for future feature re-use within the local window.

\noindent\paragraph{Global map} To clarify, we do not maintain two separate representations. Both global and local maps are made of the same keyframes and landmarks using the same inverse depth representation. The difference lies in the fact that the global map only comprises keyframes and landmarks that have been marginalized from the local map and are no longer part of the local photometric \ac{BA}. Once marginalized, the patch of pixels descriptors associated with the features are removed to minimize memory usage. However, we retain their inverse depth and variance estimates, which are held fixed until their ORB descriptor matches a feature from the active map. At this point, we propose two different mechanisms to re-use and update the global features.
\begin{itemize}
    \item Early adoption: before adding new map points to the local map, we perform feature matching between the global map and the newly added keyframe. If a match is found, the corresponding global map point is then re-activated in the local map by assigning a local patch of pixels descriptor extracted from the new keyframe. Additionally, its depth estimate and variance are initialized using the last estimates obtained before the map point was marginalized from the local map.
    \item Late fusion: during map maintenance (the update local map block in Fig. \ref{fig:proposed_architecture}), we check for matches between the currently active map points and the global ones. If a match is found, we check if the projected depth estimate of the global point is in close proximity of the local one (\ie, the local depth $- 2\sigma_d$ $\leq$ projected global depth $\leq$ local depth $+ 2\sigma_d$). If this condition is met, the global map point is re-activated and assigned to the local map point by fusing the information from their observed keyframes, as well as by combining their depth estimates and variance using \cite{engel_2013_ICCV}-Eq. (13).
\end{itemize}
Once a global map point is re-activated, its depth estimate and variance are maintained using the local photometric \ac{BA} until it is marginalized again, thereby eleminating the need for separate map maintenance processes.

\begin{figure*}[!ht]
\begin{center}
   \includegraphics[width=0.9\linewidth]{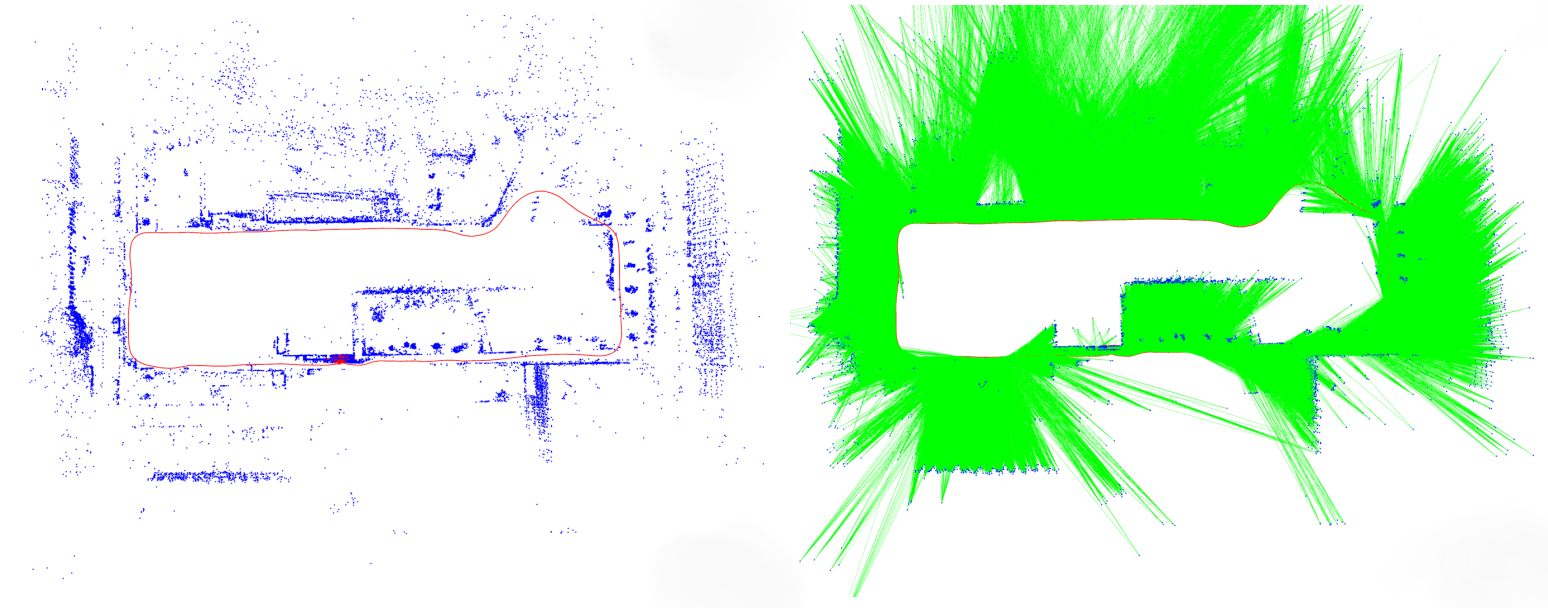}
\end{center}
   \caption{A top view sample map output from our system. (Left) The global map and traversed trajectory after loop closure and global inverse depth \ac{BA} on Sequence 49 of the TUM mono dataset \cite{mono_dataset}. (Right) The graph constraints (green lines) that were used in the full \ac{BA} to constrain the 3D map points and their corresponding keyrames in which they were observed.}
\label{fig:globalBA}
\end{figure*}

\subsection{Loop closure with hybrid graphs}
The availability of temporally connected and co-visibility information provides a valuable set of constraints that relate the keyframes and their observed landmarks. This can be leveraged to perform various types of optimizations (\textit{e.g.} \ac{PGO}, full \ac{BA}, \textit{etc.}) using various techniques (discussed in sec. \ref{sec:loopClos}).

Loop closure starts by parsing the newly added keyframe into a \ac{BoVW} model \cite{lopez_2012_TRO} to generate a global keyframe descriptor, which is then used to query the global database of keyframes for potential matches. To prevent spurious detections (which are common in LDSO \cite{Gao_2018_iros}), candidates connected to the latest keyframe in the co-visibility graph are discarded. If no loop candidates are found, which is the common case for most keyframes, the loop closure thread terminates. 
However, if a loop candidate is detected, we perform a 3D-3D map point matching between the loop candidate keyframe and the most recently added one, then use the matches to compute a corrective Sim(3) in a RANSAC implementation of \cite{horn_1987_JOSA}. The corrective Sim(3) is used to establish more 3D-3D map point matches from keyframes connected to both sides of the loop. In other words, we query the co-visibility graphs of both sides of the loop to build a set of keyframes, from which we compute more 3D-3D matches and use them to further refine the Sim(3) estimate. This typically returns a much larger number of inlier matches compared to systems like LDSO, and eventually reduces the risks of incorporating erroneous transforms into the map.
Moreover, if an insufficient number of matches are found, the loop closure thread rejects the candidate keyframe; otherwise, it uses the corrective Sim(3) to correct the poses of all keyframes that contributed feature matches from one side of the loop. Since we rely on a moving active window to explore the scene, we follow a similar approach to \cite{Gao_2018_iros} by fixing the more recent side of the loop and correcting the old observations. Aside from not breaking the priors in the local window, this has the advantage of running the loop closure correction without the need to lock the mapping thread. Thus, regular pose estimation and mapping processes can continue normally in their respective threads while the loop closure correction takes place in the third parallel thread.

The corrected keyframes are then considered fixed, and a \ac{PGO} is used to correct the remainder of the path as described in \cite{mur-artal_2015_TRO}-Eq. (9). 
The key differentiating factor between this approach and ORB-SLAM's lies in the type of constraints fed into the system. While ORB-SLAM can only use co-visibility constraints, we employ several sources. In particular, the temporally connected keyframes provide pose-pose constraints in feature deprived-environments, allowing the optimization to \textit{smooth} the path in these locations. We also include the poses of removed keyframes as part of the pose-pose constraints; these can be thought of as control points that can further help constrain the traversed trajectory.

\ac{PGO} is relatively fast to compute, requiring approximately 400 ms to correct a path made of 700 keyframes. It achieves this speed by discarding 3D observations during its path correction, which makes the results sub-optimal. To achieve optimal results, we further refine the loop closure by performing a full inverse depth \ac{BA} using all connectivity and 3D map point observations. The connected graph of the full inverse depth \ac{BA} is shown in Fig. \ref{fig:globalBA}. Note that this step is not possible in odometry methods or direct \ac{SLAM} methods like LDSO as the necessary information is not tracked. It is also not possible in previously proposed hybrid methods as they maintain different map representations for the local and global maps. 
Although the computation for this step is relatively slow, it can be executed without interfering with other thread operations. This is because it considers all keyframes from or newer than the active window at the time of loop closure detection as fixed, and only modifies marginalized keyframes and map points.

\section{Evaluation}
\label{sec:Evaluation}
We evaluate our proposed \ac{SLAM} system qualitatively and quantitatively on EuRoC \cite{euroc_2016}, KITTI \cite{geiger_2012_cvpr}, TUM mono \cite{engel2_2016_arxiv}, and TUM VI \cite{schubert_2018_tumvi} datasets. We further split the evaluation into odometry and \ac{SLAM} modes. We then assess the performance of our system in each mode of operation using applicable dataset sequences.

\subsection{Quantitative results}
\subsubsection{Trajectory error}
In this section, we evaluate the accuracy of our proposed system in both \ac{VO} and \ac{SLAM} modes against state-of-the-art methods, namely LDSO \cite{Gao_2018_iros}, DSO \cite{engel_2016_ARXIV}, ORB-SLAM3 \cite{campos_2021_tro}, and DSM \cite{Zubizarreta_2020_tro}. 
We repeat each sequence from EuRoC, KITTI, and TUM VI datasets $10$ times, and we report on the median absolute trajectory error (in meters). The results are summarized in Tab. \ref{tab:ate_perf}.

\hyphenation{Euroc}
\begin{table}[]
    \centering
    \adjustbox{width=1\textwidth}{
    \begin{tabular}{cc;{1pt/1pt}c;{1pt/1pt}c;{1pt/1pt}c;{1pt/1pt}c;{1pt/1pt}c;{1pt/1pt}c;{1pt/1pt}c;{1pt/1pt}c;{1pt/1pt}c;{1pt/1pt}c;{1pt/1pt}c}
        & Sequence & MH0l & MH02 & MH03 & MH04 & MH05 & Vl01 & Vl02 & Vl03 & V20l & V202 & V203  \\ 
        \hline
        \multirow{7}{*}{\rotcell{EuRoC}} & Ours & {\color[HTML]{FF8001} 0.035}   & {\color[HTML]{FF8001} 0.034} & 0.111                        & {\color[HTML]{FF8001} 0.111} & {\color[HTML]{1616FF} 0.064} & {\color[HTML]{FF8001} 0.039} & 0.085                        & 0.266                         & {\color[HTML]{FF8001} 0.037}  & {\color[HTML]{FF8001} 0.047}                         & 1.218                     \\
        \cdashline{2-13}[1pt/1pt] & Ours (strict VO) & 0.036                          & 0.037                        & 0.140                        & 0.334                        & 0.141                        & 0.l36                        & 0.193                        & 0.823                         & 0.051                         & 0.077                         & 1.257                        \\
        \cdashline{2-13}[1pt/1pt] & LDSO    & 0.053                          & 0.062                        & 0.114                        & 0.152                        & 0.085                        & 0.099                        & 0.087                        & 0.536                         & 0.066                         & 0.078                         & {\color[HTML]{FD0000} X}     \\
        \cdashline{2-13}[1pt/1pt] & DSO              & 0.046                          & 0.046                        & 0.172                        & 3.81                         & 0.11                         & 0.089                        & 0.107                        & 0.903                         & 0.044                         & 0.132                         & {\color[HTML]{FF8001} l.l52} \\
        \cdashline{2-13}[1pt/1pt] & ORB-SLAM3       & {\color[HTML]{1616FF} 0.016}   & {\color[HTML]{1616FF} 0.027} & {\color[HTML]{1616FF} 0.028} & 0.138                        & 0.072                        & {\color[HTML]{1616FF} 0.033} & {\color[HTML]{1616FF} 0.015} & {\color[HTML]{1616FF} 0.033*} & {\color[HTML]{1616FF} 0.023*} & {\color[HTML]{1616FF} 0.029*} & {\color[HTML]{FD0000} X}     \\
        \cdashline{2-13}[1pt/1pt] & DSM & 0.039 & 0.036                        & {\color[HTML]{FF8001} 0.055} & {\color[HTML]{1616FF} 0.057} & {\color[HTML]{FF8001} 0.067} & 0.095                        & {\color[HTML]{FF8001} 0.059} & {\color[HTML]{FF8001} 0.076}  & 0.056                         & 0.057                         & {\color[HTML]{1616FF} 0.784}
           \\ 
    \hline
    \end{tabular}
    }
    \bigskip
    \vfill
    \adjustbox{width=1\textwidth}{
    \begin{tabular}{cc;{1pt/1pt}c;{1pt/1pt}c;{1pt/1pt}c;{1pt/1pt}c;{1pt/1pt}c;{1pt/1pt}c;{1pt/1pt}c;{1pt/1pt}c;{1pt/1pt}c;{1pt/1pt}c;{1pt/1pt}c}
        & Sequence & 00	& 01 & 02 & 03 & 04 & 05 & 06 &		07	& 08 &	09 &	10  \\ 
        \hline
        \multirow{7}{*}{\rotcell{KITTI}} & Ours             & 73.12 & {\color[HTML]{1616FF} 8.65}   & {\color[HTML]{FF8001} 28.15}  & {\color[HTML]{FF8001} 1.25}  & {\color[HTML]{FF8001} 0.73}  & 118.68                       & 19.395                        & 16.425                      & {\color[HTML]{FF8001} 88.71}                         & {\color[HTML]{1616FF} 4.44}  & 11.75                       \\
        \cdashline{2-13}[1pt/1pt] & Ours (strict VO) & 104.6                        & {\color[HTML]{FF8001} 8.73}   & 97.91                         & 1.255                        & 0.735                        & 44.03                        & 57.08                         & 16.44                       & 88.94                         & 63.04                        & 11.78                       \\
        \cdashline{2-13}[1pt/1pt] & LDSO             & {\color[HTML]{FF8001} 9.322} & 11.68                         & 31.98                         & 2.85                         & 1.22                         & {\color[HTML]{1616FF} 5.1}   & {\color[HTML]{1616FF} 13.55}  & {\color[HTML]{FF8001} 2.96} & 129.02 & {\color[HTML]{330001} 21.64} & 17.36                       \\
        \cdashline{2-13}[1pt/1pt] & DSO              & 126.7                        & 165.03                        & 138.7                         & 4.77                         & 1.08                         & 49.85                        & 113.57                        & 27.99                       & 120.17 & 74.29                        & 16.32                       \\
        \cdashline{2-13}[1pt/1pt] & ORB-SLAM3       & {\color[HTML]{1616FF} 7.3}   & 519.8* & {\color[HTML]{1616FF} 25.77*} & {\color[HTML]{1616FF} 1.01}  &  1.1   & {\color[HTML]{FF8001} 6.49}  & {\color[HTML]{FF8001} 14.95}  & {\color[HTML]{1616FF} 2.79} & {\color[HTML]{1616FF} 58.6}   & {\color[HTML]{FF8001} 7.57}  & {\color[HTML]{FF8001} 8.29}                        \\
        \cdashline{2-13}[1pt/1pt] & DSM              & {\color[HTML]{FD0000} X}     & 14.27                         & {\color[HTML]{FD0000} X}      &  1.279 & {\color[HTML]{1616FF} 0.483} & 57.99 & 121.75 & 19.08                       & {\color[HTML]{FD0000} X}      & 67.87                        & {\color[HTML]{1616FF} 7.89}
           \\ 
    \hline
    \end{tabular}
    }
    \bigskip
    \vfill
    \adjustbox{width=0.7\textwidth}{
    \begin{tabular}{cc;{1pt/1pt}c;{1pt/1pt}c;{1pt/1pt}c;{1pt/1pt}c;{1pt/1pt}c;{1pt/1pt}c}
        & Sequence & Rooml &	Roorn2	& Room3	& Room4	& Room5	& Room6  \\ 
        \hline
        \multirow{7}{*}{\rotcell{TUM VI}} & Ours             & {\color[HTML]{FF8001} 0.084}  & 0.115 & {\color[HTML]{FF8001} 0.061}  & {\color[HTML]{FF8001} 0.077} & 0.084 & 0.0844 \\
        \cdashline{2-8}[1pt/1pt] & Ours (strict VO) & 0.0853 &  0.229 & 0.094  & 0.086 & {\color[HTML]{FF8001} 0.0535} & 0.0912 \\
        \cdashline{2-8}[1pt/1pt] & LDSO             & 0.128 & {\color[HTML]{FF8001} 0.086} & {\color[HTML]{FD0000} X}                             & 0.221                        & 0.114                         & {\color[HTML]{FF8001} 0.075}  \\
        \cdashline{2-8}[1pt/1pt] & DSO              &  0.177  & 0.518 & 1.364  &  0.499 & 0.107  & 0.09   \\
        \cdashline{2-8}[1pt/1pt] & ORB-SLAM3        & {\color[HTML]{1616FF} 0.042}  & {\color[HTML]{1616FF} 0.026} & {\color[HTML]{1616FF} 0.028}  & {\color[HTML]{1616FF} 0.046} & {\color[HTML]{1616FF} 0.046}  & {\color[HTML]{1616FF} 0.043}  \\
        \cdashline{2-8}[1pt/1pt] & DSM              & {\color[HTML]{FD0000} X}      & {\color[HTML]{FD0000} X} &  571.47 & {\color[HTML]{FD0000} X}     & {\color[HTML]{FD0000} X}  &  2443.5
           \\ 
    \hline
    \end{tabular}
    }
    \caption{Absolute Trajectory Error (in meters) on the EuRoC, KITTI and TUM VI datasets. Each experiment is repeated 10 times and the resulting medians are reported. Note that LDSO's, ORB-SLAM3's and DSM's results were lifted from their respective papers. Sequences marked with $\ast$ correspond to where ORB-SLAM3 is re-initializing several maps and reporting the error on the largest map. The values in blue are the most accurate, orange the second best, and red X indicates that the system did not complete.\label{tab:ate_perf}}
\end{table}
As for the TUM mono dataset \cite{engel2_2016_arxiv}, it cannot be used to evaluate SLAM systems; despite the dataset's wide ranging scenarios, it does not contain ground truth information for the entire paths, as it only provides localization information for the beginning and end of each sequence. Evaluating a system that can perform loop closure on this dataset will subsequently yield overoptimistic results that does not reflect the true performance. For this reason, we disable loop closures on this dataset and evaluate the odometry systems while enabling global map point re-use. The alignment errors are reported in Tab. \ref{Tab:rmse}

\begin{table*}[t]
    \renewcommand{\arraystretch}{1.1}
    \centering
    \adjustbox{width=\textwidth}{
    \begin{tabular}{cc;{1pt/1pt}c;{1pt/1pt}c;{1pt/1pt}c;{1pt/1pt}c;{1pt/1pt}c;{1pt/1pt}c;{1pt/1pt}c;{1pt/1pt}c;{1pt/1pt}c;{1pt/1pt}c;{1pt/1pt}c;{1pt/1pt}c;{1pt/1pt}c;{1pt/1pt}c;{1pt/1pt}c;{1pt/1pt}c;{1pt/1pt}c;{1pt/1pt}c;{1pt/1pt}c;{1pt/1pt}c;{1pt/1pt}c;{1pt/1pt}c;{1pt/1pt}c;{1pt/1pt}c;{1pt/1pt}c}
    & Sequence & 1 & 2  & 3  & 4  & 5  & 6  & 7  & 8  & 9   & 10 & 11 & 12 & 13 & 14 & 15 & 16 & \multicolumn{1}{c}{17} & 18 & 19 & 20 & 21 & 22 & 23 & 24 & 25                      \\ 
    \hline
    \multirow{4}{*}{\rotcell{TUM mono}} & Ours (VO)                        & 0.83                   & \textcolor{blue}{0.36} & \textcolor{blue}{0.82} & 0.78                   & \textcolor{blue}{1.72} & \textcolor{red}{1.07}                   & 0.52 & 0.4                    & \textcolor{blue}{0.601} & 0.32                   & \textcolor{blue}{0.47} & 0.66                   & \textcolor{blue}{1.12} & 0.86                   & \textcolor{blue}{0.67} & \textcolor{blue}{0.47} & 2.42                   & \textcolor{blue}{1.37} & 2.35                   & \textcolor{blue}{0.56} & \textcolor{blue}{2.75} & \textcolor{blue}{3.15} & 0.55                   & 0.36                   & \textcolor{blue}{0.51}  \\
    \cdashline{2-27}[1pt/1pt] & DSO                              & \textcolor{blue}{0.53} & 0.57                   & \textcolor{red}{3.39}                   & \textcolor{blue}{0.69} & 1.74                   & \textcolor{blue}{0.81} & \textcolor{red}{0.6}                    & \textcolor{blue}{0.35} & \textcolor{red}{0.62}                    & \textcolor{blue}{0.29} & 0.59                   & \textcolor{blue}{0.6}  & 1.39                   & \textcolor{blue}{0.72} & 0.83                   & 0.51                   & \textcolor{blue}{2.3}  & 1.58                   & \textcolor{blue}{1.9}  & \textcolor{red}{0.75}                   & 4.22                   & 3.95                   & \textcolor{blue}{0.48} & \textcolor{blue}{0.3}  & 0.82                    \\
    \cdashline{2-27}[1pt/1pt] & ORB-SLAM3 (VO) & \textcolor{red}{0.93}  & \textcolor{red}{51.99} & 0.83  & \textcolor{red}{2.44}  & \textcolor{red}{X}     & 1.02  & \textcolor{blue}{0.36}  & \textcolor{red}{178.9} & 0.604  & \textcolor{red}{0.73}  & \textcolor{red}{1.26}  & \textcolor{red}{1.17}  & \textcolor{red}{2.33}  & \textcolor{red}{6.37}  & \textcolor{red}{1.45}  & \textcolor{red}{0.91}  & \textcolor{red}{8.71}  & \textcolor{red}{X}     & \textcolor{red}{3.51}  & 0.7   & \textcolor{red}{X}     & \textcolor{red}{X}     & \textcolor{red}{3.66}  & \textcolor{red}{0.53}  & \textcolor{red}{1.11}   \\ 
    \hline\hline
    & Sequence                         & 26                     & 27                     & 28                     & 29                     & 30                     & 31                     & 32                     & 33                     & 34                      & 35                     & 36                     & 37                     & 38                     & 39                     & 40                     & 41                     & 42                     & 43                     & 44                     & 45                     & 46                     & 47                     & 48                     & 49                     & 50                      \\ 
    \hline
    \multirow{4}{*}{\rotcell{TUM mono}} & Ours (VO)                        & \textcolor{blue}{1.35} & \textcolor{red}{1.59}  & \textcolor{blue}{1.23} & \textcolor{blue}{0.32} & \textcolor{red}{1.05}  & 0.67                   & 0.49                   & \textcolor{blue}{1.19} & \textcolor{blue}{0.56}  & 0.68                   & 1.79                   & 0.39                   & 0.9                    & 1.4                    & \textcolor{blue}{1.72} & \textcolor{blue}{0.35} & \textcolor{blue}{0.79} & \textcolor{blue}{0.35} & \textcolor{blue}{0.28} & \textcolor{blue}{0.77} & 0.75                   & \textcolor{blue}{1.44} & \textcolor{blue}{0.81} & \textcolor{blue}{0.68} & \textcolor{blue}{0.53}   \\
    \cdashline{2-27}[1pt/1pt] & DSO                              & 3.34                   & \textcolor{blue}{0.97} & 2.19                   & 0.43                   & \textcolor{blue}{0.66} & \textcolor{blue}{0.59} & \textcolor{blue}{0.32} & 1.45                   & 0.88                    & \textcolor{blue}{0.58} & \textcolor{red}{5.85}  & \textcolor{blue}{0.38} & \textcolor{blue}{0.77} & \textcolor{blue}{1.29} & 1.81                   & 0.9                    & 0.89                   & 0.46                   & 0.55                   & 1.26                   & \textcolor{blue}{0.61} & 1.52                   & 1.09                   & \textcolor{red}{X}     & 0.77                    \\
    \cdashline{2-27}[1pt/1pt] & ORB-SLAM3 (VO)                  & \textcolor{red}{13.8}  & 1.49                   & \textcolor{red}{13.49} & \textcolor{red}{6.38}  & 0.78                   & \textcolor{red}{X}     & \textcolor{red}{1.85}  & \textcolor{red}{2.06}  & \textcolor{red}{1.8}    & \textcolor{red}{9.02}  & \textcolor{blue}{1.23} & \textcolor{red}{0.5}   & \textcolor{red}{9.37}  & \textcolor{red}{18.25} & \textcolor{red}{X}     & \textcolor{red}{X}     & \textcolor{red}{10.12} & \textcolor{red}{1.13}  & \textcolor{red}{1}     & \textcolor{red}{3.17}  & \textcolor{red}{7.43}  & \textcolor{red}{9.43}  & \textcolor{red}{3.78}  & 5.66 & \textcolor{red}{3.7}    \\
    \hline
    \end{tabular}
    }
    \caption{Alignment error on the TUM mono dataset \cite{engel2_2016_arxiv}. The evaluated systems are run in \ac{VO} mode (\textit{i.e.} loop closure is disabled) while allowing global map point re-use for both ORB-SLAM3 and ours. The blue and red values represent the lowest and highest errors respectively. Note that LDSO does not support such functionality and reduces to DSO in such mode.\label{Tab:rmse}}
\end{table*}

\subsubsection{Computational cost}
\label{sec:compute_cost}
The computational cost and memory footprint analysis was performed on the same CPU Intel core i9-8950 @ 2.9GHz with 32 GB RAM; no GPU acceleration was used. To gauge tractability of the SLAM systems we consider the following metrics:

\begin{itemize}
    \item{Tracking time (ms): the time required for the system to process an input frame and generate the corresponding camera pose as output. Tracking is performed every frame.}
    \item {Mapping time (ms): the time for the mapping thread to finish integrating a new keyframe into the map. Mapping is run whenever a new keyframe is added.}
    \item {Loop closure time (ms): the duration required to perform loop closure error estimation and correction. This process runs fairly infrequently, \ie, whenever a loop closure is detected. }
    \item {Memory cost: the amount of memory (MB) consumed by the system per keyframe.\footnote{We don't report on the total memory cost as it can vary significantly  across datasets and sequences. Instead, we rely on a metric that can be consolidated over several sequences.}}
    \item{Loop closure found: The number of independent sequences per dataset in which at least $1$ loop closure is detected.}
    
\end{itemize}

Similarly, we run each sequence $10$ times and compute the median of the stats per sequence; we then report on the mean performance of each system across all different sequences per dataset. The results are shown in Tab. \ref{tab:UFVSComputeCost}. 
Note that our system, LDSO, and ORB-SLAM3 use a \ac{BoVW} dictionary to detect loop closure candidates. The dictionary must be loaded in memory and has a constant size (not included in the table). In contrast, DSM does not perform loop closure, and therefore does not require the dictionary; however, we don't report on the numbers for DSM in the table as its performance was sub-realtime on the majority of the datasets we tested and failed on a significant amount of sequences outside EuRoC. For those interested, on the datasets where it actually succeeded, the average tracking time was 10 ms, the mapping time 660 ms and the memory consumed was about 9.26 MB per keyframe. 

\hyphenation{Euroc}
\begin{table*}[t]
\centering
\renewcommand{\arraystretch}{1.3}
\adjustbox{width=0.8\textwidth}{
\begin{tabular}{cc;{1pt/1pt}c;{1pt/1pt}c;{1pt/1pt}c;{1pt/1pt}c;{1pt/1pt}c}
                                 &    Metric        & Tracking time(ms)       & Mapping time (ms)        & Loop closure time (ms)   & Memory (MB/KF)             & Loop closure found  \\ 
\hline
\multirow{4}{*}{\rotcell{EuRoC}} & Ours       & \textcolor{blue}{12.43} & \textcolor{blue}{73.06}  & 2041                      & \textcolor{blue}{0.712} & 3                   \\ 
\cdashline{2-7}[1pt/1pt]
                                 & LDSO       & 13.49                   & 130.38                   & \textcolor{blue}{104.48} & 0.97                      & 10                  \\ 
\cdashline{2-7}[1pt/1pt]
                                 & ORB-SLAM3 & \textcolor{red}{17.4}   & \textcolor{red}{241.63}  & \textcolor{red}{3074}    & \textcolor{red}{1.465}  & 5                   \\ 
\hline\hline
\multirow{4}{*}{\rotcell{KITTI}} & Ours       & 13.76                   & \textcolor{blue}{74.25}  & 5488.2                    & \textcolor{blue}{0.64}  & 5                   \\ 
\cdashline{2-7}[1pt/1pt]
                                 & LDSO       & \textcolor{blue}{8.98}  & 109.66                   & \textcolor{blue}{475.73} & 0.796               
                                 & 11                  \\ 
\cdashline{2-7}[1pt/1pt]
                                 & ORB-SLAM3 & \textcolor{red}{23.87}  & \textcolor{red}{130.82}  & \textcolor{red}{7952}    & \textcolor{red}{1.08}   & 6                   \\ 
\hline\hline
\multirow{4}{*}{\rotcell{TUM VI}} & Ours       & 13.38                   & \textcolor{blue}{79.267} & 669.3                    & 0.85                     & 3                   \\ 
\cdashline{2-7}[1pt/1pt]
                                 & LDSO       & \textcolor{blue}{9.46}  & 114.73                   & \textcolor{blue}{84.1}   & \textcolor{blue}{0.58}   & 6                   \\ 
\cdashline{2-7}[1pt/1pt]
                                 & ORB-SLAM3 & \textcolor{red}{23.87}  & \textcolor{red}{318.68}  & \textcolor{red}{813.5}   & \textcolor{red}{2.38}  & 2                   \\
\hline
\end{tabular}
}
\caption{Processing stats for the various SLAM systems. Each sequence in each dataset is repeated $10$ times and we report on the overall dataset average of the sequence stats medians. The blue and red values represent the best and worst scores respectively.}
\label{tab:UFVSComputeCost}
\end{table*}

\subsection{Qualitative Results}
\begin{figure}[!htbp]
\begin{center}
   \includegraphics[height=0.5\textheight,keepaspectratio]{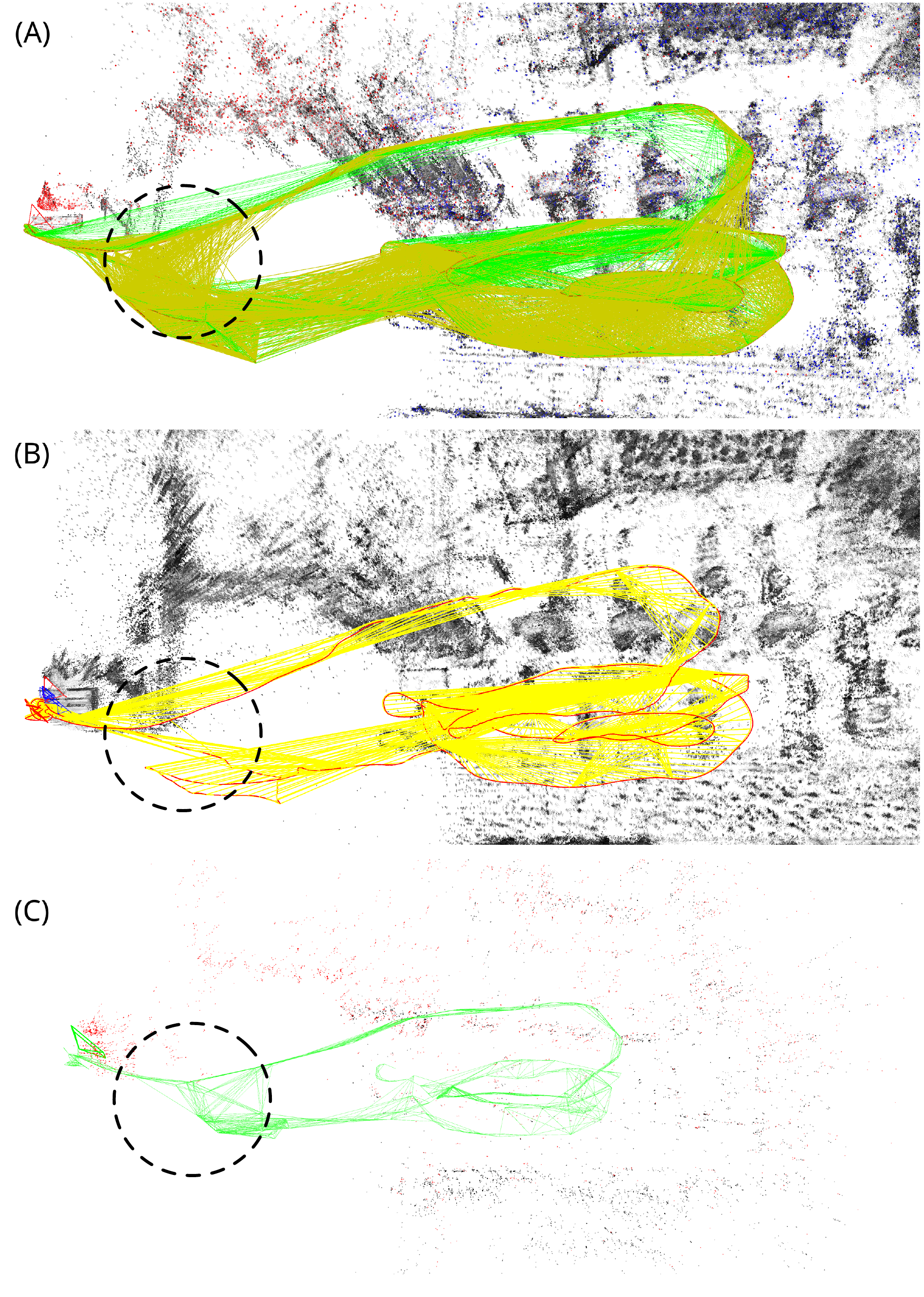}
\end{center}
\caption[Constraints available for \ac{PGO} in various VSLAM systems.]{Constraints available for \ac{PGO} in the MH\_01\_easy (left images) sequence from the EuRoC \cite{euroc_2016} dataset. Each colored line represents a constraint between 2 keyframes. (A) shows both the pose-pose (green) and co-visibility (dark yellow) constraints of our proposed approach. (B) shows the pose-pose constraints from LDSO \cite{Gao_2018_iros}. (C) shows the co-visibility constraints from ORB-SLAM2 \cite{Mur-Artal_Tardos_2017}. Note how the methods that use co-visibility constraints (A) and (C) can re-establish correspondences between old and new measurements when loop closure is detected (circled areas where the density of constraints increase between the keyframes along both ends of the loop), whereas methods that use pose-pose constraints only (B) cannot recover such constraints, as they only track temporal constraints even after loop closure.}
\label{fig:UFVS-graphs}
\end{figure}

Fig. \ref{fig:UFVS-graphs} shows the constraints between the keyframes available for optimization. We plot our proposed hybrid graph \textit{vs} LDSO's pose-pose graph \textit{vs} ORB-SLAM's co-visibility graph. The use of both co-visibility and pose-pose graphs allows our system to generate a relatively dense network of constraints when compared to ORB-SLAM or LDSO (as shown in Fig. \ref{fig:UFVS-graphs}). This is because the hybrid co-visibility graph keeps track of both temporally connected keyframes and those that are connected by common observations. The former allows the system to establish a prior on keyframe motion in feature-deprived environments, while the latter allows it to establish connections based on re-observed features, which can be especially helpful after performing loop closure (encircled areas in Fig. \ref{fig:UFVS-graphs}). This allows us to recognize and add new connections between keyframes that were once unconnected due to large drift, and accordingly re-use global points in the local odometry window post a loop closure event. 
On the other hand, LDSO's pose-pose constraints cannot be updated to account for new constraints when loop closure takes place, and global map points cannot be re-used. 

Note that ORB-SLAM employs a keyframe culling strategy by removing redundant keyframes from its map, resulting in a fairly pruned set of co-visibility constraints. We employ a similar strategy, except we keep the culled keyframes to act as temporal pose-pose constraints.

\section{Discussion}
\subsection{Processing time}
The average tracking time per frame for our system is 13 ms, during which an indirect optimization first takes place, followed by a joint multi-objective optimization. In contrast, LDSO being a direct system, only requires around 10.6 ms to perform direct image alignment. On the other hand, ORB-SLAM3 needs an average of 21.7 ms to extract features from multiple pyramid levels and compute the frame's pose. 

The entire mapping process of our approach requires 75 ms per keyframe on average to generate and maintain both the direct and indirect global maps. 
In comparison, the mapping thread of LDSO requires 117 ms to process a keyframe, while ORB-SLAM3 requires 230 ms. Note that ORB-SLAM's mapping process performs a local \ac{BA} every time a new keyframe is added; while this results in improved accuracy, it comes at the cost of increased mapping time, requiring almost three times as much as our system to maintain its global map. 
Finally DSM requires about 660 ms per keyframe. This significant increase in computational cost is mainly due to the use of a photometric global map that  DSM attempts to query when a new keyframe is added. Additionally, DSM performs pyramidal photometric \ac{BA} on three levels, contributing to the higher time requirements. In contrast, our method, along with LDSO, perform the local photometric \ac{BA} on a single pyramid level. It's worth noting that our system's speed in maintaining both local and global maps is comparable to pure odometry methods like DSO, which only maintains a local map at a cost of 62 ms per keyframe.

As for loop closure, LDSO is typically the fastest as it only performs a \ac{PGO}, whereas ORB-SLAM3 and our system perform a full \ac{BA}. However, since this process runs on a third parallel thread, the longer processing time does not significantly impact the overall \ac{SLAM}'s performance. This is especially true in our case since we adopt a similar strategy to LDSO, \ie, we update the old trajectory and keep the latest keyframes fixed in order to maintain coherence in the local photometric optimization priors. 

\subsection{Memory Cost}
We employ several strategies to keep our memory consumption low.
By maintaining a co-visbility graph, it is possible to apply a pruning strategy similar to that used in ORB-SLAM. We remove redundant keyframes from the global map that have a high degree of overlap (more than 90\%) with other keyframes in terms of shared map points. However, unlike ORB-SLAM, our system will still utilize pruned keyframes in the pose-pose constraints of our hybrid graph, which can provide better constraints to the overall trajectory for long range loop closures. Moreover, the use of descriptor sharing directly translates to reduced memory cost as we do not need to keep track of separate landmark information for local-global representations. Compared to our proposed approach, LDSO consumes about 6\% more memory (MB) per keyframe, despite not maintaining a global reusable 3D map. ORB-SLAM3, on the other hand, consumes 120\% more. These numbers highlight the significant impact of descriptor sharing and the use of a single representation for both local and global maps in reducing memory overhead in our system.

\subsection{Trajectory accuracy}
\label{sec:traj_acc}

Our proposed \ac{SLAM} system achieves better performance than LDSO on most sequences, even though both systems use a local moving window to explore the scene. 
The performance improvement is attributed to three reasons: (1) the improved accuracy of the hybrid pose estimation process, (2) the improved connectivity graph that contains both co-visibility and pose-pose constraints, and (3) the global \ac{BA} that optimizes the global map.
On the other hand, our system scores between ORB-SLAM3 and DSM, outperforming them on few sequences, while being a close second candidate on others.

The mixed results can be attributed to several factors, the most prominent one being the small sized environments of the datasets themselves: all the sequences of EuRoC and TUM VI that have ground truth poses take place in small rooms. In such scenarios, ORB-SLAM's local \ac{BA} encompasses the entire map and behaves like a global \ac{BA}, optimizing over the total map for every keyframe added. This, however, results in a significantly increased computational cost per keyframe, requiring more than 240 ms in EuRoC and more than 300 ms in TUM VI. Whereas in exploratory datasets like KITTI, the computational cost drops to 130 ms, and so does their trajectory performance.

In contrast, our system maintains a strictly fixed seven-keyframe window for performing local \ac{BA}. A global \ac{BA} is only triggered at loop closure, which happens fairly infrequently in small rooms as we identify and re-use previously observed global map points without loop closure. This is shown in Tab. \ref{tab:UFVSComputeCost}, where loop closure is only triggered in 3 out of the 11 sequences of the EuRoC dataset. In the remaining sequences, full \ac{BA} is never performed and our system operates as a pure odometry system with the ability to re-use global points. The impact of this ability is contrasted when comparing the results of our system in strictly \ac{VO} mode, where the use of global points is disabled (shown in Tab. \ref{tab:ate_perf} under "ours (strict \ac{VO})").

Another reason for the performance difference in exploratory sequences could be the lack of photometric calibration in the KITTI dataset, which puts our system at a disadvantage. The impact of this issue is mitigated in the TUM mono dataset where we evaluate our system in odometry mode (no loop closure) but with global map points re-use enabled, and compare it to the same setup in ORB-SLAM3 and DSO. The results shown in Tab.  \ref{Tab:rmse} demonstrate that our system outperforms both of them on the majority of the sequences.

\subsection{Limitations and further improvements}
While the proposed system can theoretically perform at least as good as ORB-SLAM3, our current implementation under-performs in small room-like environments (reasons discussed in \ref{sec:traj_acc}). The underlying problem in our current implementation is that we don't re-activate previously marginalized keyframes. Instead, we add new keyframes and re-activate old map points within them. This was done to cater to the needs of the local moving window photometric \ac{BA}, where we aim to refresh the global map points using recent photometric pixel patches. DSM, on the other hand, addresses this issue by re-activating previously marginalized keyframes, not just map points, through a pyramidal photometric \ac{BA} that can handle large baselines and intensity changes. However, this is computationally very expensive and makes their system sub-realtime, which contradicts our overall goal of achieving high performance at low computational cost.
In our system, we have observed the potential of using the presence of indirect residuals in the map to enhance the resilience of the local photometric \ac{BA} to large baselines and intensity changes. Similar to what we have done in joint multi-objective tracking, leveraging these indirect residuals could enable us to re-activate old keyframes. This remains an open problem that we plan to address in our future work. 

\section{Conclusion}
\label{sec:Conclusion}
In this work, we have demonstrated the advantages of combining direct and indirect formulations through a descriptor sharing approach. This integration enables us to create a unified representation for both local and global maps, introducing several key capabilities that are typically absent in individual frameworks or other hybrid methods. These capabilities include the ability to perform global Bundle Adjustment on a shared map representation, to re-activate previously observed map points, to perform keyframe culling, and to extract and maintain hybrid connectivity graphs. 
By incorporating both temporally connected and co-visibility constraints, our approach allows for loop closure using concurrent Pose Graph Optimization over both types of constraints. Additionally, the sub-optimal results of the \ac{PGO} are further refined in a global inverse depth \ac{BA}, a process not typically feasible in other hybrid methods. We have validated the system's capabilities through its computational and memory efficiency, showcasing competitive performance compared to other methods on most tested sequences. 

\section*{Acknowledgment}
\begin{sloppypar}
The authors would like to thank the Ontario Centres of Excellence (OCE), the University Research Board (URB) at the American University of Beirut, and the Natural Sciences and Engineering Research Council of Canada (NSERC) for their support in conducting this research. The work was further supported by the Didymos Horizon Europe project under grant number 101092875–DIDYMOS-XR.
\end{sloppypar}



\bibliographystyle{elsarticle-num} 
  \bibliography{younes_bib}





\end{document}